\documentclass{article}

\usepackage[final, nonatbib]{neurips_2024}

\usepackage[numbers]{natbib}

\usepackage{amsmath}

\usepackage[utf8]{inputenc} 
\usepackage[T1]{fontenc}    
\usepackage{url}            
\usepackage{booktabs}       
\usepackage{amsfonts}       
\usepackage{nicefrac}       
\usepackage{microtype}      
\usepackage{xcolor}         
\usepackage{mathtools}
\usepackage{float}

\usepackage{wrapfig}

\usepackage{subcaption}

\usepackage{pgfplots}
\usepgfplotslibrary{colorbrewer}
\pgfplotsset{compat = 1.15, cycle list/Set1-8}
\usetikzlibrary{pgfplots.statistics, pgfplots.colorbrewer}
\usepackage{pgfplotstable}

\usepackage{hyperref}       

\newcommand{\inlineSubsection}[2]{
  \refstepcounter{subsection} 
  \noindent\textbf{\thesubsection\ #1}\label{#2}
}

\title{Continuous Autoregressive Models with Noise Augmentation Avoid Error Accumulation}

\author{
  Marco Pasini$^1$\thanks{This work is supported by the EPSRC UKRI Centre for Doctoral Training in Artificial Intelligence and Music (EP/S022694/1) and Sony Computer Science Laboratories Paris.}
  \And
  Javier Nistal$^2$
  \And
  Stefan Lattner$^2$
  \And
  Gy\"orgy Fazekas$^1$
  \AND\\$^1$Queen Mary University, London, UK\\
  $^2$Sony Computer Science Laboratories, Paris, France
}

\begin{document}

\maketitle

\vspace{-6mm}
\begin{abstract}
\vspace{-2mm}
    Autoregressive models are typically applied to sequences of discrete tokens, but recent research indicates that generating sequences of continuous embeddings in an autoregressive manner is also feasible. However, such Continuous Autoregressive Models (CAMs) can suffer from a decline in generation quality over extended sequences due to error accumulation during inference. We introduce a novel method to address this issue by injecting random noise into the input embeddings during training. This procedure makes the model robust against varying error levels at inference. We further reduce error accumulation through an inference procedure that introduces low-level noise. Experiments on musical audio generation show that CAM substantially outperforms existing autoregressive and non-autoregressive approaches while preserving audio quality over extended sequences. This work paves the way for generating continuous embeddings in a purely autoregressive setting, opening new possibilities for real-time and interactive generative applications.
\end{abstract}
\vspace{-4mm}

\section{Introduction}
\vspace{-3mm}
\begin{wrapfigure}{r}{0.5\textwidth}
\vspace{-1.cm}
\centering
\includegraphics[trim={0cm 0cm 0 0.14cm},clip,width=0.5\textwidth]{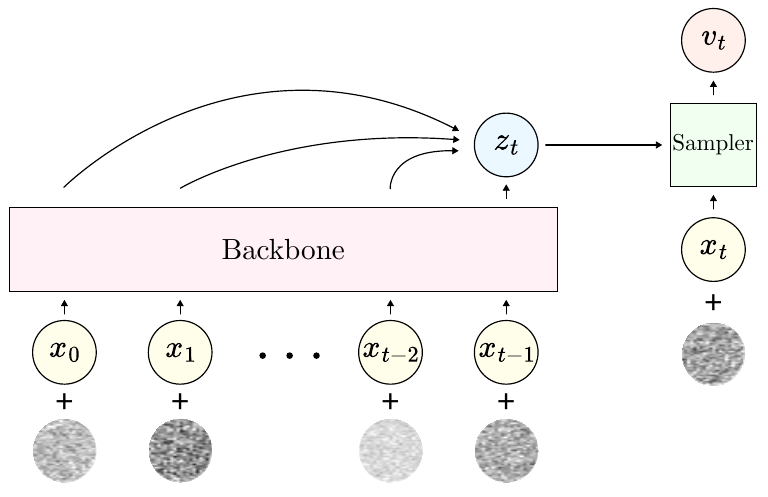}
\vspace{-0.5cm}
\caption{Training process of CAM. The causal Backbone receives as input a sequence of continuous embeddings with noise augmentation. It outputs $z_t$, which is used by the Sampler as conditioning to denoise a noise-corrupted version of $x_t$.}
\label{fig:training}
\end{wrapfigure}

Autoregressive Models (AMs) have become ubiquitous in various domains, achieving remarkable success in natural language processing tasks~\cite{gpt2, gpt3}. These models operate by predicting the next element in a sequence based on the preceding elements, a principle that lends itself naturally to inherently sequential data like text. However, their application to continuous data, such as images and audio waveforms, presents unique challenges.

First, autoregressive models for image and audio generation have traditionally relied on discretizing data into a finite set of tokens using techniques like Vector Quantized Variational Autoencoders (VQ-VAEs) \cite{vqvae, fsq}. This discretization allows models to operate within a discrete probability space, enabling the use of the cross-entropy loss, in analogy to their application in language models. However, quantization methods typically require additional losses (e.g., commitment and codebook losses) during VAE training and may introduce a hyperparameter overhead. Secondly, continuous embeddings can encode information more efficiently than discrete tokens (the same information can be encoded in shorter sequences), enabling AMs to perform faster inference than their discrete counterparts. Recent works explore training autoregressive models on continuous embeddings \cite{mar, givt}, bypassing the need for quantisation. While promising, these methods are particularly sensitive to error accumulation during inference which produces a distribution shift, hindering the generation quality when using a sequentially autoregressive approach (GPT-style). Instead, these works rely on cumbersome non-sequential masking schemes (e.g., predicting embeddings at random positions at each step) \cite{mar} and careful tuning of training and inference-time techniques \cite{givt} to indirectly tackle error accumulation. These techniques not only add complexity but also impede the exploitation of efficient inference techniques developed in the context of Large Language Models (LLMs) for discrete tokens (e.g., key-value cache \cite{kvcache}), potentially preventing their adoption by a wider research community.

In this work, we introduce a simple yet intuitive method to counteract error accumulation and reliably train purely autoregressive models on ordered sequences of continuous embeddings without complexity overhead. As shown in Fig. \ref{fig:training}, by augmenting the training data with random data-noise mixtures, we encourage the model to learn to distinguish between real and ``erroneous'' signals, making it robust to error propagation during inference. Additionally, we introduce a simple inference technique that involves adding a small amount of artificial noise to the generated embeddings, further increasing resilience to accumulated errors. We refer to models trained using the proposed technique as CAMs (Continuous Autoregressive Models). We demonstrate the effectiveness of CAM through unconditional generation experiments on an audio dataset of music stems, since we believe that fast GPT-style models in the audio and music domains could unlock powerful interactive applications, such as real-time music accompaniment systems and end-to-end speech conversational models. Our results show that CAM substantially outperforms existing autoregressive and non-autoregressive baselines regarding generation quality. Moreover, CAM does not demonstrate any degradation when generating longer sequences, indicating its effectiveness in mitigating error accumulation.
CAM unlocks the potential of autoregressive models for efficient and interactive generation tasks, opening new possibilities for real-time applications.

\vspace{-3mm}
\section{Related Work}\label{sec:related_work}
\vspace{-3mm}
Autoregressive models have achieved remarkable success in natural language processing, becoming the dominant approach for tasks like language modeling \cite{transformer, gpt, gpt2, gpt3}.
Extending autoregressive models to image and audio generation has been an active area of research. Early attempts directly model the raw data, as exemplified by PixelRNN \cite{pixelrnn} and WaveNet \cite{wavenet}, which operate on sequences of quantized pixels and audio samples, respectively. However, these approaches are computationally demanding, particularly for high-resolution images and long audio sequences.
To address this challenge, recent works have shifted towards modeling compressed representations of images and audio, typically obtained using autoencoders. A popular approach involves discretizing these representations using Vector Quantized Variational Autoencoders (VQ-VAEs) \cite{vqvae}, enabling autoregressive models to operate on a sequence of discrete tokens.
This strategy has led to significant advances in both image~\cite{vqgan, maskgit} and audio generation \cite{jukebox, copet_simple_2023}.

Recent approaches explore training AMs directly on continuous embeddings. GIVT \cite{givt} uses the AM's output to parameterise a Gaussian Mixture Model (GMM), enabling training with cross-entropy loss. At inference, continuous embeddings can be sampled directly from the GMM. Despite its success in high-fidelity image generation, GIVT requires additional techniques, such as variance scaling and normalizing flow adapters, that add complexity to the model and training procedure.
Alternative approaches like Masked Autoregressive models (MAR) \cite{mar} learn the per-token probability distribution using a diffusion procedure. A shallow MLP is used to sample a continuous embedding conditioned on the output of an autoregressive transformer. However, the authors show that a sequential autoregressive model with causal attention (i.e., GPT-style \cite{gpt}) performs poorly in this setting and requires bidirectional attention and random masking strategies during training. 
Our work tackles this inconvenience to make training of GPT-style models feasible, which we believe can unlock new avenues for real-time interactive applications, especially in the field of audio generation.

\vspace{-3mm}
\section{Background}\label{sec:background}
\vspace{-3mm}
\inlineSubsection{Denoising Diffusion Models (DDMs)}{sec:ddm_background} are a class of generative models that learn a given data distribution $p(x)$ by gradually corrupting it with noise (\emph{diffusion}) and then learning to reverse this process (\emph{denoising}). Specifically, they model the score function of the noise-perturbed data distribution at various noise levels.
Given a set of noise levels ${\sigma_t}_{t=1}^T$, we can define a series of perturbed data distributions $p_{\sigma_t}(x_t) = \int p(x) \mathcal{N}(x_t ; x, \sigma^2_t I) d x$.
For each noise level $\sigma_t$ with $t=0,1,...,T$, DDMs learn a score $s_{\theta}(x,t)$ approximating that of the corresponding perturbed distribution:
$s_{\theta}(x, t) \approx \nabla_{x} \log p_{\sigma_t}(x)$ 
where $s_{\theta}$ is typically implemented as a neural network, $x$ is the input data point, and $t$ is the noise level.
The training objective is then to minimize the weighted sum of Fisher Divergences between the model and the true score functions at all noise levels:
\begin{equation}
\mathcal{L} = \sum_{t=1}^T \lambda(t) \mathbb{E}_{p_{\sigma_t}\left(x_t\right)}\left[\left|s_{\theta}(x, t)-\nabla_{x} \log p_{\sigma_t}\left(x_t\right)\right|_2^2\right],
\end{equation}
where $\lambda(t)$ is a positive weighting function that depends on the noise level.
Once trained, DDMs generate new samples using annealed Langevin dynamics: starting from a Gaussian random sample, the process iteratively refines the sample by following the direction of the score function at decreasing noise levels, eventually arriving at a clean sample from the target distribution $p(x)$.
\vspace{-1mm}

\inlineSubsection{Rectified Flow (RF)}{sec:rf_background}~\cite{rectifiedflow} offers a conceptually simpler and more general alternative to DDMs and was shown to perform better than competing diffusion frameworks on latent embedding generation tasks \cite{sd3}. RF directly connects two arbitrary distributions $\pi_0$ and $\pi_1$ by following straight line paths. In the basic framework, $\pi_0$ is the data distribution, and $\pi_1$ is the noise distribution, typically sampled from a standard Gaussian.  Given a set of samples $(x_0 \sim \pi_0, x_1 \sim \pi_1)$, a rectified flow is defined by the ordinary differential equation (ODE) $dz_t = v(z_t, t)dt$,
where $z_t$ represents the data point at time $t$, and $v(z_t, t)$ is the so-called \emph{drift force} and it is parameterized by a neural network trained to minimize the loss:
\begin{equation}
    \mathcal{L} = \mathbb{E} \left[ || (x_1 - x_0) - v(t x_1 + (1-t)x_0, t) ||^2 \right].
\end{equation}
This objective encourages the flow to follow the straight line paths connecting $x_0$ and $x_1$, resulting in a more efficient deterministic mapping than other diffusion-based frameworks.
\vspace{-1mm}

\inlineSubsection{Autoregressive Models for Continuous Embeddings}{sec:mar_background}, as proposed in MAR~\cite{mar}, employ diffusion models to predict the next element $x_t$ in a sequence, based on the preceding elements $(x_0, x_1, ..., x_{t-1})$. This can be formulated as estimating the conditional probability $p(x_t | x_0, x_1, ..., x_{t-1})$.\footnote{Note that, in this case, $x_t$ indicates the element of the $(T+1)$-long data sequence at position $t$.} To predict $x_t$ MAR first transforms $\left( x_0,...,x_{t-1} \right)$ into a vector $z_t$ using a \textit{Backbone} neural network, and then model $p(x_t | z_t)$ using a diffusion process. A second network, \textit{Sampler}, predicts a noise estimate from $y_t$, which represents $x_t$ corrupted with noise $\varepsilon \sim \mathcal{N}(0, I)$. The training objective is formulated as:
\begin{equation}
    \mathcal{L} = \mathbb{E}_{t} \left[ \| \varepsilon - \text{Sampler}(y_t | z_t) \|^2 \right] \quad \text{where} \quad z_t = \text{Backbone}\left( x_0,...,x_{t-1} \right).
\end{equation}
This objective encourages the model to learn to denoise the corrupted embedding $y_t$ and recover the original $x_t$ based on the information about previous timesteps contained in the condition $z_t$.
At inference time, the model generates a new sequence by iteratively predicting conditioning vectors $z_t$ based on the previously generated elements and then using a reverse diffusion process to sample $x_t$ from the learned distribution $p(x_t | z_t)$.
MAR, however, shows that naive training of GPT-style models—using causal modeling of ordered sequences—fails to deliver compelling results. Instead, masked modeling and bidirectional attention mechanisms are necessary to achieve performance on par with non-autoregressive approaches. We argue that masked modeling, which involves predicting random timesteps, mitigates error accumulation by discouraging the model from relying exclusively on preceding time steps to generate the current one.

\vspace{-3mm}
\section{Proposed Method}\label{sec:method}
\vspace{-3mm}
\textbf{Training}
As seen in Sec.~\ref{sec:mar_background}, while MAR \cite{mar} enables training AMs on continuous embeddings, a significant challenge emerges when generating ordered sequences: error accumulation. At inference, prediction errors propagate throughout the generation process and compound at each subsequent predicted time step, leading to a divergence from the learned data distribution. To address this, we introduce a novel strategy that injects noise during training to simulate erroneous predictions, encouraging the model to be robust against it (see Fig. \ref{fig:training}). Specifically, we assume that at inference, the \textit{Sampler} (see Sec.~\ref{sec:mar_background}) generates embeddings that can be expressed as a linear combination of the real data $x_t \sim \pi_0$ and an error $\varepsilon \sim \mathcal{N}(0,I)$, weighted by an unknown error level $k_t$:
\begin{equation}
    \tilde{x}_t = k_t*\varepsilon + (1-k_t)*x_t. 
\end{equation}
We can then simulate inference conditions during training, aligning the distribution of embeddings with those generated during inference, which inherently exhibit error accumulation. This can help us mitigate the effects of the distribution shift.
Specifically, our solution involves sampling $k_t \sim \mathcal{U}(0,1)$ for each timestep during training and to feed noise-perturbed sequences $(\tilde{x}_0, \tilde{x}_1, ..., \tilde{x}_T)$ to the \textit{Backbone}. Importantly, and differently from the noise level in DDMs, we do not explicitly inform the \textit{Backbone} about error levels $k_t$. This results in the backbone being trained as a discriminative model, which must distinguish between real and ``error'' signals for each timestep in its input to provide the most informative condition $z_t$ to the \textit{Sampler}. Performing this noise augmentation strategy at training time allows us to simulate the error accumulation effect during inference for any error level in $(0,1)$. 
As for the \textit{Sampler}, we use the RF framework (see Sec.~\ref{sec:rf_background}) in tandem with AMs for continuous embeddings as explained in \ref{sec:mar_background}. Given $y_t = \sigma_t*\varepsilon + (1-\sigma_t)*x_t$, and a noise level $\sigma$ sampled from a lognormal distribution with $m=0$ and $s=1$~\cite{sd3},
the objective function of the end-to-end system can be expressed as:
\begin{equation}
    \mathcal{L} = \mathbb{E}_{t} \left[ \| v_t - \text{Sampler}(y_t |\sigma_t, z_t) \|^2 \right] \quad \text{with} \quad z_t = \text{Backbone}\left( \tilde{x}_0,...,\tilde{x}_{t-1} \right),
\end{equation}
where $v_t = x_t - \varepsilon$ is the drift.
During training, we drop out $z_t$ 20\% of the time and substitute it with a learnable embedding $z_{\text{SOS}}$. At inference, following GPT-style models, we prompt the \textit{Sampler} with the start-of-sentence (SOS) embedding $z_{\text{SOS}}$ to sample the first element of the generated sequence.
\vspace{-1mm}

\textbf{Inference}
At inference, CAM generates a new sequence of embeddings autoregressively, following the temporal order of the sequence. Given the initial conditioning vector $z_{\text{SOS}}$, the \textit{Sampler} generates the first embedding $\hat{x}_1$ by performing an iterative reverse diffusion process (see Sec.~\ref{sec:ddm_background}). 
Subsequent embeddings are generated by concatenating $\hat{x}_{t-1}$ to the existing sequence of previously generated embeddings. 
The sequence is fed as input to the \textit{Backbone} to produce the conditioning vector $z_t$, which is then used by the \textit{Sampler} to generate $\hat{x}_t$. This process is repeated iteratively until the desired sequence length is reached. Since the \textit{Sampler} is parameterised by a shallow MLP, the computation required by the denoising process can be negligible compared to the forward pass of the \textit{Backbone}.
To further dampen the effects of error accumulation, we observe that adding a small constant amount of Gaussian noise $k_{\text{inf}}$ to each generated embedding $\hat{x}_t$ before feeding it back to the \textit{Backbone} can yield higher quality when generating long sequences. We hypothesize that this noise helps to reduce the mismatch between the Gaussian distribution used for perturbation during training and the actual distribution of errors of the \textit{Sampler}'s predictions.

\vspace{-3mm}
\section{Experiments and Results}\label{sec:results}
\vspace{-3mm}
\textbf{Datasets:} For training and evaluation purposes, we use an internal dataset composed of $\sim20,000$ single-instrument recordings covering various instruments and musical styles. Each audio file is stereo and has a 48 kHz sample rate. We preprocess the dataset by extracting continuous latent representations using an in-house stereo version of Music2Latent \cite{music2latent}, a state-of-the-art audio autoencoder. This results in compressed latent embeddings with a sampling rate of $\sim12$ Hz and a dimensionality of $64$. During training, we randomly crop each embedding sequence to 128 frames, corresponding to approximately $10$ seconds of stereo audio.
\vspace{-1mm}

\textbf{Implementation Details:} The \textit{Backbone} in CAM is a transformer with a pre-LN configuration, $16$ layers, $\textit{dim}=768$, $\textit{mlp\_mult}=4$, $\textit{num\_heads}=4$. We use absolute learned positional embeddings. The \textit{Sampler} is an MLP with $8$ layers, $\textit{dim}=768$, $\textit{mlp\_mult}=4$. Both $z_t$ and $y_t$ are concatenated and fed as input to the MLP, while information about the noise level $\sigma_t$ is introduced via AdaLN \cite{dit}. The total number of parameters for the entire model is ~150 million.
Regarding training, we use AdamW \cite{adamw} with $\beta_1=0.9$, $\beta_2=0.999$, weight decay $=0.01$, and a learning rate of $1e-4$. All models are trained for 400k iterations with a batch size of 128.
\vspace{-1mm}

\textbf{Baselines:} We compare CAM against several autoregressive and non-autoregressive baselines: \textit{GIVT} models \cite{givt} with 8 and 32 modes, the model proposed by \cite{mar} in its fully autoregressive and causal configuration (we denote this model as \textit{MAR}), and a non-autoregressive diffusion model trained using the \textit{Rectified Flow} \cite{rectifiedflow} framework.
We also provide the results of \textit{MAR} trained using Rectified Flow instead of its original linear noise-prediction objective and of \textit{GIVT} trained using our proposed noise augmentation technique.
To ensure a fair comparison in model capacity, we use the same architecture for all models, and we increase the number of transformer layers to $21$ in those models that do not use a \textit{Sampler} to roughly match the total number of parameters. We provide audio samples at \href{https://sonycslparis.github.io/cam-companion/}{sonycslparis.github.io/cam-companion/}.
\vspace{-1mm}

\textbf{Evaluation Metrics:} We use Frechet Audio Distance (FAD) \cite{fad} to evaluate the quality of generated samples. We use FAD calculated using CLAP features \cite{clap}, which accepts 10-second high-sample rate samples as input and has been shown to exhibit a stronger correlation with perceived quality compared to VGGish features \cite{fad_correlation}. FAD is calculated using a reference set of 10,000 samples and background sets of 1,000 samples, and we report the average over 5 evaluations. All samples are 10 seconds long. To evaluate the influence of error accumulation, we also use $\text{FAD}_{\text{acc}}$, which is the FAD obtained by the 10 seconds of audio that are autoregressively generated after the first 10 seconds.
\vspace{-2mm}



\pgfplotsset{compat=1.17}
\pgfplotsset{/pgfplots/custom legend/.style={legend image code/.code={\draw [only marks,mark=square*]
plot coordinates {(0.cm,0cm)};}},}

\begin{figure}[h]
    \vspace{-.5cm}
    \centering
    \begin{subfigure}[t]{0.5\textwidth}
        \vspace{-2.5cm}
        \captionsetup{margin={-6cm,0cm}}
        \caption{}
        \vspace{-0.5cm}
        \begin{center}
            \begin{tabular}{lcc}
                \toprule
                \textbf{Model} & \textbf{FAD} & $\textbf{FAD}_{\text{acc}}$ \\
                \midrule
                MAR & 0.453 & 0.458 \\
                MAR RF & \textbf{0.442} & \textbf{0.453} \\
                \bottomrule
            \end{tabular}
        \end{center}
        \label{tab:rectified}
        \vspace{0.2cm}
        \begin{minipage}[h]{\textwidth}
            \includegraphics[width=\textwidth]{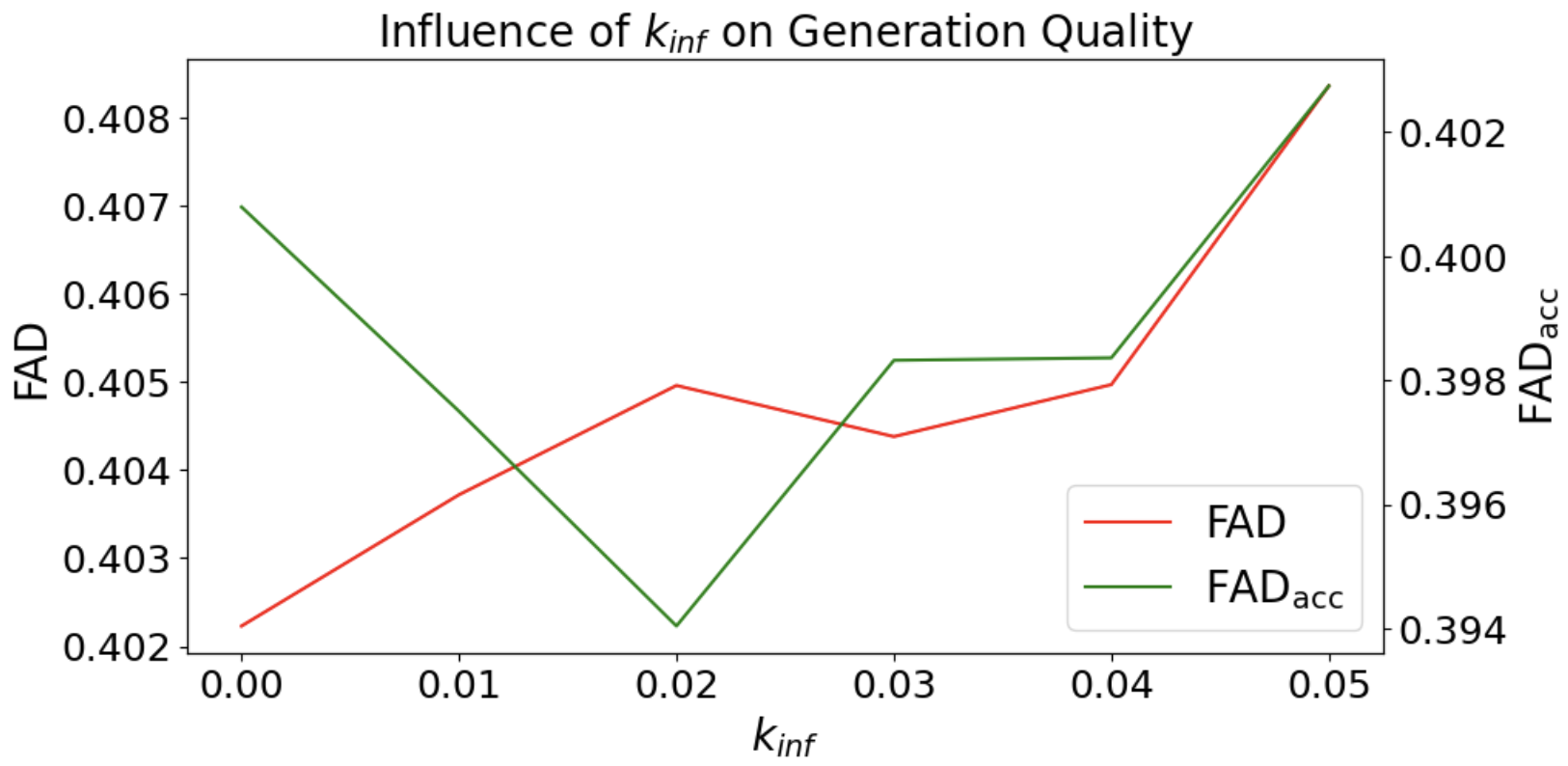}
            \captionsetup{margin={-6cm,0cm}}
            \vspace{-0.8cm}
            \caption{}
            \label{fig:inference_noise}
        \end{minipage}
    \end{subfigure}
    \hspace{0.1cm}
    \begin{subfigure}[h]{0.45\textwidth}
        \vspace{1cm}
        \begin{tabular}{lcc}
            \toprule
            \textbf{Model} & \textbf{FAD} & $\textbf{FAD}_{\text{acc}}$ \\
            \midrule
            \textbf{Non-Autoregressive} & & \\
            Rectified Flow & 0.448 & n/a \\
            \midrule
            \textbf{Autoregressive} & & \\
            GIVT (8 modes) & 0.889 & 0.950 \\
            GIVT (32 modes) & 0.865 & 0.931 \\
            GIVT+noise (32 modes) & 0.514 & 0.511 \\
            MAR RF & 0.442 & 0.453 \\
            CAM (Ours) & \textbf{0.405} & \textbf{0.394} \\
            \bottomrule
        \end{tabular}
        \caption{}
        \label{tab:baseline_comparison}
    \end{subfigure}
    \caption{(a) Comparison between \textit{MAR} trained using noise-prediction with linear schedule and \textit{MAR RF} using Rectified Flow. (b) Influence of $k_{inf}$ on FAD and $\text{FAD}_{\text{acc}}$. (c) Comparison of CAM with Autoregressive and Non-Autoregressive Baselines.}

\end{figure}

\vspace{-3mm}
\textbf{Influence of Rectified Flow}:
In \ref{tab:rectified}, we first compare \textit{MAR} trained using the original noise-prediction with linear schedule diffusion framework to the same model trained using a Rectified Flow formulation. For each model, we use the number of denoising steps in the range (10,100) that results in the lowest FAD. The model trained using Rectified Flow achieves a lower FAD.
\vspace{-1mm}

\textbf{Influence of Inference Noise}:
We evaluate FAD and $\text{FAD}_{\text{acc}}$ when CAM uses different values of $k_{\text{inf}}$ in the $[0, 0.05]$ range. Fig. \ref{fig:inference_noise} shows the results obtained for each noise level.
Remarkably, we note how with $k=0.02$, $\text{FAD}_{\text{acc}} < \text{FAD}$, pointing to an improvement in generation quality for longer generations. A possible explanation of this result is: since the \textit{Backbone} receives a maximum context of $\sim 10$ seconds, it generates all embeddings after the $10$ seconds mark using a full context, which may result in higher quality embeddings.
We use $k=0.02$ for all subsequent experiments.
\vspace{-1mm}

\textbf{Comparison with Baselines}:
We evaluate CAM and the baselines concerning their ability to generate high-fidelity audio. The $\text{FAD}_{\text{acc}}$ metric directly evaluates the resilience of the models to error accumulation. A model that does not suffer from error accumulation would achieve the same results on both the first and the second 10-second generated audio sequence. Since we are not interested in evaluating or minimizing inference speed, for each model relying on diffusion sampling we use the number of denoising steps in the range (10,100) that results in the lowest FAD. We also use variance scaling for \textit{GIVT} to sample embeddings with a temperature of $t=0.9$, which we empirically find to result in a lower FAD. A technique to simulate sampling with different temperatures has also been proposed for \textit{MAR} \cite{mar}; however, we find that the best metrics are obtained with $t=1$.

As we show in Tab.~\ref{tab:baseline_comparison}, CAM outperforms all autoregressive and non-autoregressive baselines on FAD metrics. CAM also exhibits a decrease in FAD when autoregressively generating longer sequences. The same result can be noticed for \textit{GIVT} when trained with our proposed noise augmentation, which also performs vastly better than the original \textit{GIVT} models. This demonstrates that our proposed training approach can be successfully adapted to different categories of autoregressive models for continuous embeddings. In contrast, all other autoregressive baselines show a degradation in audio quality as the generated sequence length increases.

\vspace{-4mm}
\section{Conclusion}\label{sec:conclusion}
\vspace{-4mm}
This paper introduced CAM, a novel method for training purely autoregressive models on continuous embeddings that directly addresses the challenge of error accumulation. By introducing random noise into the input embeddings during training, we force the model to learn robust representations resilient to error propagation. Additionally, a carefully calibrated noise injection technique employed during inference further mitigates error accumulation. Our experiments demonstrate that CAM substantially outperforms existing autoregressive and non-autoregressive models for audio generation, achieving the lowest FAD while maintaining consistent audio quality even when generating extended sequences. This work paves the way for new possibilities in real-time and interactive audio applications that benefit from the efficiency and sequential nature of autoregressive models.

\bibliographystyle{unsrt}
\bibliography{refs}

\begin{thebibliography}{10}

\bibitem{gpt2}
Alec Radford, Jeff Wu, et~al.
\newblock Language models are unsupervised multitask learners, 2019.

\bibitem{gpt3}
Tom~B. Brown, Benjamin Mann, Nick Ryder, Melanie Subbiah, Jared Kaplan, Prafulla Dhariwal, Arvind Neelakantan, Pranav Shyam, Girish Sastry, Amanda Askell, Sandhini Agarwal, Ariel Herbert{-}Voss, Gretchen Krueger, Tom Henighan, Rewon Child, Aditya Ramesh, Daniel~M. Ziegler, Jeffrey Wu, Clemens Winter, Christopher Hesse, Mark Chen, Eric Sigler, Mateusz Litwin, Scott Gray, Benjamin Chess, Jack Clark, Christopher Berner, Sam McCandlish, Alec Radford, Ilya Sutskever, and Dario Amodei.
\newblock Language models are few-shot learners.
\newblock In Hugo Larochelle, Marc'Aurelio Ranzato, Raia Hadsell, Maria{-}Florina Balcan, and Hsuan{-}Tien Lin, editors, {\em Advances in Neural Information Processing Systems 33: Annual Conference on Neural Information Processing Systems 2020, NeurIPS 2020, December 6-12, 2020, virtual}, 2020.

\bibitem{vqvae}
A{\"{a}}ron van~den Oord, Oriol Vinyals, et~al.
\newblock Neural discrete representation learning.
\newblock In {\em Advances in Neural Information Processing Systems 30}, December 2017.

\bibitem{fsq}
Fabian Mentzer, David Minnen, Eirikur Agustsson, and Michael Tschannen.
\newblock Finite scalar quantization: {VQ-VAE} made simple.
\newblock In {\em The Twelfth International Conference on Learning Representations, {ICLR} 2024, Vienna, Austria, May 7-11, 2024}. OpenReview.net, 2024.

\bibitem{mar}
Tianhong Li, Yonglong Tian, He~Li, Mingyang Deng, and Kaiming He.
\newblock Autoregressive image generation without vector quantization.
\newblock {\em arXiv preprint arXiv:2406.11838}, 2024.

\bibitem{givt}
Michael Tschannen, Cian Eastwood, and Fabian Mentzer.
\newblock Givt: Generative infinite-vocabulary transformers.
\newblock {\em arXiv preprint arXiv:2312.02116}, 2023.

\bibitem{kvcache}
Noam Shazeer.
\newblock Fast transformer decoding: One write-head is all you need.
\newblock {\em arXiv preprint arXiv:1911.02150}, 2019.

\bibitem{transformer}
Ashish Vaswani, Noam Shazeer, et~al.
\newblock Attention is all you need.
\newblock In {\em Advances in Neural Information Processing Systems 30}, December 2017.

\bibitem{gpt}
Alec Radford and Karthik Narasimhan.
\newblock Improving language understanding by generative pre-training, 2018.

\bibitem{pixelrnn}
A{\"{a}}ron van~den Oord, Nal Kalchbrenner, and Koray Kavukcuoglu.
\newblock Pixel recurrent neural networks.
\newblock In Maria{-}Florina Balcan and Kilian~Q. Weinberger, editors, {\em Proceedings of the 33nd International Conference on Machine Learning, {ICML} 2016, New York City, NY, USA, June 19-24, 2016}, volume~48 of {\em {JMLR} Workshop and Conference Proceedings}, pages 1747--1756. JMLR.org, 2016.

\bibitem{wavenet}
A{\"{a}}ron van~den Oord, Sander Dieleman, et~al.
\newblock {WaveNet}: {A} generative model for raw audio.
\newblock In {\em The 9th {ISCA} Speech Synthesis Workshop}, September 2016.

\bibitem{vqgan}
Patrick Esser, Robin Rombach, et~al.
\newblock Taming transformers for high-resolution image synthesis.
\newblock In {\em {IEEE} Conference on Computer Vision and Pattern Recognition (CVPR)}, June 2021.

\bibitem{maskgit}
Huiwen Chang, Han Zhang, et~al.
\newblock Maskgit: Masked generative image transformer.
\newblock In {\em {IEEE/CVF} Conference on Computer Vision and Pattern Recognition, {CVPR} 2022, New Orleans, LA, USA, June 18-24, 2022}, 2022.

\bibitem{jukebox}
Prafulla Dhariwal, Heewoo Jun, et~al.
\newblock Jukebox: A generative model for music.
\newblock {\em arXiv preprint arXiv:2005.00341}, 2020.

\bibitem{copet_simple_2023}
Jade Copet, Felix Kreuk, et~al.
\newblock Simple and {Controllable} {Music} {Generation}, June 2023.
\newblock arXiv:2306.05284 [cs, eess].

\bibitem{rectifiedflow}
Xingchao Liu, Chengyue Gong, and Qiang Liu.
\newblock Flow straight and fast: Learning to generate and transfer data with rectified flow.
\newblock In {\em The Eleventh International Conference on Learning Representations, {ICLR} 2023, Kigali, Rwanda, May 1-5, 2023}. OpenReview.net, 2023.

\bibitem{sd3}
Patrick Esser, Sumith Kulal, et~al.
\newblock Scaling rectified flow transformers for high-resolution image synthesis.
\newblock {\em arXiv preprint arXiv:2403.03206}, 2024.

\bibitem{music2latent}
Marco Pasini, Stefan Lattner, and George Fazekas.
\newblock Music2latent: Consistency autoencoders for latent audio compression.
\newblock {\em arXiv preprint arXiv:2408.06500}, 2024.

\bibitem{dit}
William Peebles and Saining Xie.
\newblock Scalable diffusion models with transformers.
\newblock In {\em {IEEE/CVF} International Conference on Computer Vision, {ICCV} 2023, Paris, France, October 1-6, 2023}, 2023.

\bibitem{adamw}
Ilya Loshchilov and Frank Hutter.
\newblock Decoupled weight decay regularization.
\newblock In {\em 7th International Conference on Learning Representations, {ICLR} 2019, New Orleans, LA, USA, May 6-9, 2019}. OpenReview.net, 2019.

\bibitem{fad}
Kevin Kilgour, Mauricio Zuluaga, et~al.
\newblock Fr{\'{e}}chet audio distance: {A} reference-free metric for evaluating music enhancement algorithms.
\newblock In {\em 20th Annual Conference of the International Speech Communication Association (INTERSPEECH)}, September 2019.

\bibitem{clap}
Yusong Wu, Ke~Chen, et~al.
\newblock Large-scale contrastive language-audio pretraining with feature fusion and keyword-to-caption augmentation.
\newblock In {\em {IEEE} International Conference on Acoustics, Speech and Signal Processing {ICASSP} 2023, Rhodes Island, Greece, June 4-10, 2023}, 2023.

\bibitem{fad_correlation}
Modan Tailleur, Junwon Lee, et~al.
\newblock Correlation of fr$\backslash$'echet audio distance with human perception of environmental audio is embedding dependant.
\newblock {\em arXiv preprint arXiv:2403.17508}, 2024.

\end{thebibliography}

\end{document}